\documentclass[english]{lni}

\usepackage[latin1]{inputenc}

\usepackage{pgf-pie}  
\usepackage{subfig}
\usepackage{graphicx}
\usepackage{fancyhdr}
\usepackage{listings} 
\usepackage{changepage} 
\usepackage[figurename=Fig., tablename=Tab.]{caption}[2008/04/01]
\usepackage[rootbib]{chapterbib}

\renewcommand{\headrulewidth}{0.4pt} 
\makeatletter
\def\blfootnote{\xdef\@thefnmark{}\@footnotetext}
\makeatother

\author{Mustafa Ekrem Erak{\i}n\textsuperscript{1\textsuperscript{\textasteriskcentered}}, U\u{g}ur Demir\textsuperscript{1\textsuperscript{\textasteriskcentered}}, Haz{\i}m Kemal Ekenel\textsuperscript{1} }

\title{On Recognizing Occluded Faces in the Wild}
\begin{document}

\begingroup\renewcommand\thefootnote{1}
\footnotetext{Faculty of Computer and Informatics, Dept. of Computer Engineering, 34469 Maslak, Istanbul, TURKEY, erakin20@itu.edu.tr,  demirug16@itu.edu.tr,  ekenel@itu.edu.tr}
\endgroup

\begingroup\renewcommand\thefootnote{\textasteriskcentered}
\footnotetext{Equal contribution}
\endgroup

\maketitle

\renewcommand{\refname}{References}

\pagestyle{fancy}
\fancyhead{} 
\fancyhead[RO]{\small On Recognizing Occluded Faces in the Wild \hspace{25pt}  \hspace{0.05cm}}
\fancyhead[LE]{\hspace{0.05cm}\small  \hspace{25pt} Mustafa Ekrem Erak{\i}n, U\u{g}ur Demir, Haz{\i}m Kemal Ekenel}

\fancyfoot{} 

\setcounter{footnote}{1} 
\renewcommand{\headrulewidth}{0.4pt} 

\begin{abstract}
Facial appearance variations due to occlusion has been one of the main challenges for face recognition systems. To facilitate further research in this area, it is necessary and important to have occluded face datasets collected from real-world, as synthetically generated occluded faces cannot represent the nature of the problem. In this paper, we present the Real World Occluded Faces (ROF) dataset, that contains faces with both upper face occlusion, due to sunglasses, and lower face occlusion, due to masks. We propose two evaluation protocols for this dataset. Benchmark experiments on the dataset have shown that no matter how powerful the deep face representation models are, their performance degrades significantly when they are tested on real-world occluded faces. It is observed that the  performance drop is far less when the models are tested on synthetically generated occluded faces. The ROF dataset and the associated evaluation protocols are publicly available at the following link https://github.com/ekremerakin/RealWorldOccludedFaces.



\end{abstract}

\section{Introduction}
With the recent advancements in deep learning and its application to computer vision problems, state-of-the-art face recognition systems have achieved excellent results on various datasets, such as LFW \cite{LFWTech}, AgeDB-30 \cite{moschoglou2017agedb}, and MegaFace \cite{kemelmachershlizerman2015megaface}. 
As the performance on these well-known datasets
converges, researchers started to divert their attention towards more challenging problems. One of these challenges is recognizing occluded faces in the wild \cite{zeng2020survey}. To catalyze further research on this topic, in this paper, we present the Real World Occluded Faces (ROF) dataset, that contains faces with both upper face and lower face occlusions. To test the authenticity of the dataset, we participated in a masked face recognition challenge \cite{9484337}. Our model, fine-tuned on real life masked images, outperformed models trained on larger, synthetically generated masked face training sets, leading to the best performance among 16 other academic submissions \cite{9484337}. 





There have been several works that studied the effects of many different appearance variations on face recognition performance \cite{mehdipour2016comprehensive}, \cite{karahan2016image},  \cite{grm2018strengths}. In this paper, we will be addressing specifically the occlusion problem using a real-world occluded face dataset. \cite{mehdipour2016comprehensive} used AR face dataset \cite{martinez1998ar} that contains occluded face images collected in a constrained environment, while \cite{karahan2016image} and \cite{grm2018strengths} used synthetic occlusions built on top of LFW \cite{LFWTech}. Our experiments show that real world occlusions are more challenging than their synthetic counterparts.

Previous studies show that deep CNN based face recognition models trained on VGGFace \cite{VGGface}, faces major performance drops when confronted by sunglasses and scarves. \cite{mehdipour2016comprehensive} reports the performance with sunglasses occlusion in the range of 30-35\% on the AR face dataset \cite{martinez1998ar}, which is a 110 identity face image dataset that is collected in a controlled environment with cooperating subjects, a rather easy benchmark for a modern face recognition model. Another study \cite{karahan2016image} uses synthetic occlusions to test the face recognition performance again using a VGGFace pre-trained model \cite{VGGface}. Occluded face images are generated by applying black boxes on samples from the LFW dataset \cite{LFWTech}. Different occlusion types are simulated by applying these black boxes in different locations. The study reports 25.94\% face recognition accuracy against sunglasses effect. 

The main contributions of this paper can be summarized as follows: (i) We introduced an in-the-wild occlusion dataset for face recognition, 
(ii) we proposed two evaluation protocols and analyzed the impact of upper face and lower face occlusion on face recognition performance, 
(iii) we show that real-world face occlusion poses a more challenging problem for face recognition systems. We also visualized the results and discuss the outcomes in detail. 
\vspace{-5 mm}
\section{The Dataset}


Real World Occluded Faces (ROF) dataset contains face images with real-life upper-face and lower-face occlusions, due to sunglasses and face masks, respectively. The dataset consists of 6421 neutral face images, 4627 face images with sunglasses, and 678 face images with masks. There are 47 subjects with neutral, masked, and sunglasses images, 114 subjects with neutral and sunglasses images, while 20 subjects have only neutral and masked images. The identities are collected from a list of celebrities and politicians. All of the images are from real-life scenarios and contain large variations in terms of pose and illumination. The images were downloaded from Google Image Search using the pipeline described in VGGFace2 study \cite{cao2018vggface2}. On average there are 50 neutral images, 30 sunglasses images and 15 masked images per identity.   

\textbf{Dataset Collection}

Dataset collection is done using a modified version of the pipeline described in VGGFace2 \cite{cao2018vggface2}. A name list consisting of public figures, i.e., politicians, celebrities, sports players, etc., were collected. For every name in the list 100 images were downloaded for each type of face image we are after. A reference image was extracted from the collected neutral images for every name using the image size and face count within the image to try to get the best possible reference. Duplicates were removed using perceptual hashing and faces were detected and cropped from the remaining images using a combination of RetinaFace \cite{deng2019retinaface} and MTCNN \cite{xiang2017joint}. 

\begin{figure}[htb]
\centering
\resizebox{\columnwidth}{!}{\includegraphics{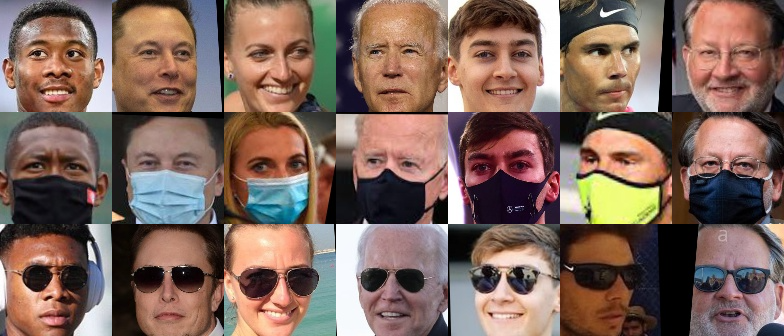}}
\caption{Samples of neutral, masked, and sunglasses images for the same subjects from the ROF dataset}
\label{fig:datasetsamples}
\end{figure}

Then using the reference images and a ResNet50 \cite{he2015deep} trained on VGGFace2 \cite{cao2018vggface2}, face embeddings were extracted for every remaining image and compared with the subject's respective reference image's embedding, using cosine distance as the similarity metric. For neutral images, candidate images with a similarity above 0.5 were selected while for occluded images, the threshold was set to 0.2. Finally, filtered face images were manually verified and stored. Overall, manual work was limited and the bottleneck was finding the appropriate identities that would have both sunglasses and face mask images, which proved to be a niche category. Figure \ref{fig:datasetsamples} shows sample images from the ROF dataset. \\

\vspace{-5mm}
\section{Deep Face Models}
We utilized three different deep learning architectures to examine the performance degradation when encountered with facial occlusions, namely ArcFace \cite{deng2019arcface}, VGGFace2 \cite{cao2018vggface2}, and MobilFaceNet \cite{chen2018mobilefacenets}. 

ArcFace \cite{deng2019arcface} is a state-of-the-art face recognition model that achieved excellent performance on various face recognition datasets, such as LFW \cite{LFWTech}, AgeDB-30 \cite{moschoglou2017agedb}, and MegaFace \cite{kemelmachershlizerman2015megaface}. ArcFace is trained on MS1MV2 \cite{deng2019arcface}, which is a revised version of the MS-Celeb-1M dataset \cite{guo2016msceleb1m}. MS1MV2 contains 85,000 identities and 5.8 million images. In this work, we used three different ArcFace architectures to represent various model complexities. Different Arcface architectures are denoted as Arcface-N which corresponds to a ResNet-N model pre-trained on MS1MV2 dataset.


VGGFace2 is a large-scale dataset containing 9131 identities and 3.3 million samples. Researchers used the dataset to train deep learning models, and it was one of the state-of-the-arts. We used their ResNet-50 pre-trained model throughout our experiments \cite{cao2018vggface2}.

Since ArcFace and VGGFace2 mainly use ResNet as the backbone architecture, their model complexities are not suitable for mobile devices. Therefore, we tested the model also on MobilFaceNet \cite{chen2018mobilefacenets} to analyze the performance degradation of a smaller model. MobilFaceNet used in this work is trained on the MS1MV2 dataset \cite{deng2019arcface}.

Throughout our experiments, we employed 512-dimensional feature embeddings. For distance metrics, Euclidean distance is utilized for ArcFace and MobilFaceNet, and cosine similarity is used for VGGFace2. ArcFace and MobilFaceNet pre-trained models are adopted from the Insightface repository\footnote{https://github.com/deepinsight/insightface}. VGGFace2 pre-trained model is adopted from the verified VGGFace2 repository.
\vspace{-7mm}

\section{Experimental Setup}

\vspace{-3mm}
In this section, we present the experimental setups to evaluate the occlusion robustness of the deep face recognition methods. We also analyze and compare the differences between the effects of synthetically crafted and real-world occluded face images. We present two experiment protocols. The first protocol investigates the effects of upper face occlusions, while the second one assesses the performance against lower face occlusions. Both protocols also probe with synthetic occlusions and compare the results with the ones obtained on the real world occluded samples. The image and identity counts across protocols are given in Table \ref{tab:protocols}.

\begin{table}[htb]
\centering
\makebox[\textwidth]{%
\begin{tabular}{l l l l l l}
\hline
Protocol & Identities & Gallery & Synthetic Probe & Sunglasses & Masked \\
\hline
1   &   161 &   483 &   5322    &   4627&   -\\
2   &   67  &   199 &   1800    &   - &   464\\
\hline
\end{tabular}}
\caption{Total number of identities and images for each protocol}
\label{tab:protocols}
\end{table}

\vspace{-3mm}

\begin{figure}
\centering
\begin{tabular}{cccccc}
\subfloat[Neutral Image]{\includegraphics[width = 0.9in]{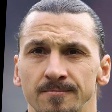}} &
\subfloat[Wearing Sunglasses]{\includegraphics[width = 0.9in]{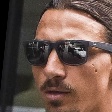}} &
\subfloat[Wearing Mask]{\includegraphics[width = 0.9in]{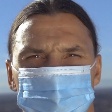}}\\
\subfloat[Upper-face Occlusion]{\includegraphics[width = 0.9in]{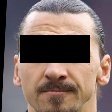}}&
\subfloat[Lower-face Occlusion]{\includegraphics[width = 0.9in]{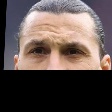}} &
\subfloat[Synthetic Mask]{\includegraphics[width = 0.9in]{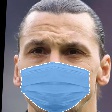}} &
\end{tabular}
\caption{a) Non-occluded face image, b) Upper face occlusion due to wearing sunglasses, c) Lower face occlusion due to wearing a mask, c) Synthetic upper face occlusion, d) Synthetic lower face occlusion, e) Synthetic mask generation for lower face occlusion}
\label{fig:occlusiontypes}
\end{figure}

For data preprocessing, CosFace \cite{wang2018cosface} and SphereFace \cite{liu2018sphereface} papers are followed. First, the bounding box and five facial landmarks, namely, eyes, nose, and mouth corners, are obtained using MTCNN \cite{Zhang_2016}. Afterwards, similarity transform is applied to images for face normalization. Then, images are cropped and resized to $112\times112$.


For testing against upper-face occlusions, we used the ROF sunglasses dataset in the first protocol. We also generated synthetic fixed upper-face occlusions that cover the eye region. Real sunglasses and synthetic occlusions can be seen in Figure \ref{fig:occlusiontypes}b and \ref{fig:occlusiontypes}d, respectively.

In the second protocol, for lower-face occlusions we used the ROF mask dataset. Synthetic lower face occlusions are generated by fixing the nose and mouth area and using the mask generator \cite{anwar2020masked} published in a recent study. Samples from real and synthetic lower face occlusions can be seen in Figure \ref{fig:occlusiontypes}c and \ref{fig:occlusiontypes}f, respectively.

\vspace{-5mm}

\section{Experimental Results and Discussion}

In this section we present the experimental results. We performed both face identification and verification experiments and assessed the effect of occlusion in both scenarios.
\vspace{-4mm}
\subsection{Impact of Occlusions on Face Identification}

\textbf{Protocol 1:} The experimental results are presented in Table \ref{tab:protocol1rec} for all used deep face models. Each row corresponds to the obtained correct classification rates on a specific probe set. Samples from probe sets are shown in Figure \ref{fig:occlusiontypes}. All models are found to be very successful when classifying face images that do not contain occlusion, as can be seen from the first row. Arcface is found to be more robust compared to MobileFaceNet and VGGFace2, when a part of the face is occluded synthetically, either by painting the corresponding region with black or generating an artificial mask. However, even Arcface's performance deteriorates when it is tested on real world occluded faces that contain sunglasses.



\begin{table}[]
\centering
\makebox[\textwidth]{%
\begin{tabular}{cccccc}
\hline
\multicolumn{1}{l}{\textbf{}} & \textbf{Arcface-100} & \textbf{Arcface-50} & \textbf{Arcface-34} & \textbf{MobilFaceNet} & \textbf{VGGFace2} \\ \hline
\textbf{Occlusion Type}       & \textbf{Top 1}       & \textbf{Top 1}      & \textbf{Top 1}      & \textbf{Top 1}        & \textbf{Top 1}    \\
\hline
\textbf{No occlusion}         & 99.57\%              & 99.34\%             & 99.17\%             & 98.89\%               & 98.12\%           \\
\textbf{Wearing sunglasses}   & 86.60\%              & 84.18\%             & 83.51\%             & 77.16\%               & 76.83\%           \\
\textbf{Upper occlusion}      & 98.25\%              & 95.92\%             & 95.43\%             & 83.13\%               & 75.65\%           \\
\textbf{Lower occlusion}      & 98.21\%              & 96.81\%             & 96.64\%             & 86.98\%               & 88.56\%           \\
\textbf{Synthetic masked}     & 98.53\%              & 97.16\%             & 96.56\%             & 89.57\%               & 89.59\%           \\ \hline
\end{tabular}}
\caption{Face identification results using protocol 1 (Arcface-N denotes the ResNet architecture with N layers, for VGGFace2 ResNet50 was used)}
\label{tab:protocol1rec}
\end{table}

\textbf{Protocol 2:} In Table \ref{tab:protocol2rec}, we present the experimental results using protocol 2. The outcomes are similar to the ones obtained using protocol 1. The deep face models are found to be very successful when there is no occlusion in the probe images. Arcface is found to be superior to MobileFaceNet and VGGFace2, when a part of the face is occluded synthetically, either by painting the corresponding region with black or generating an artificial mask. However, again, even Arcface's performance deteriorates when it is tested on real world occluded faces that contain masks.

These results show that synthetically generated occlusions do not reflect the nature of the real-world occlusions. One reason could be due to the fact that the synthetic occlusions contain the same texture and covers the same regions across different faces. However, real world occlusions contain different textures and cover different parts of the faces depending on the style of the sunglasses or the type of the mask and the way the person wears it.

\begin{figure}
\centering
\begin{tabular}{cccccc}
\subfloat[Neutral Image]{\includegraphics[width = 0.9in]{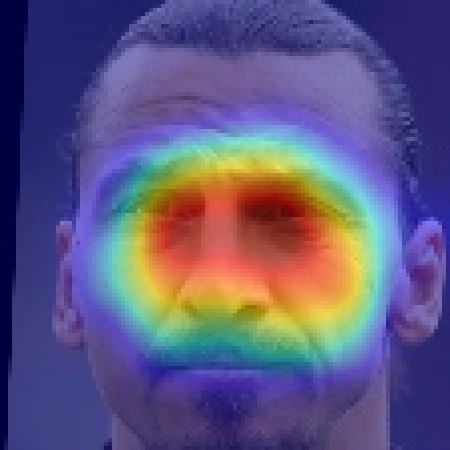}} &
\subfloat[Wearing Sunglasses]{\includegraphics[width = 0.9in]{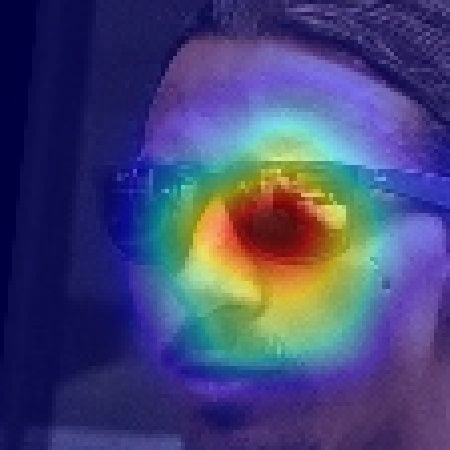}} &
\subfloat[Wearing Mask]{\includegraphics[width = 0.9in]{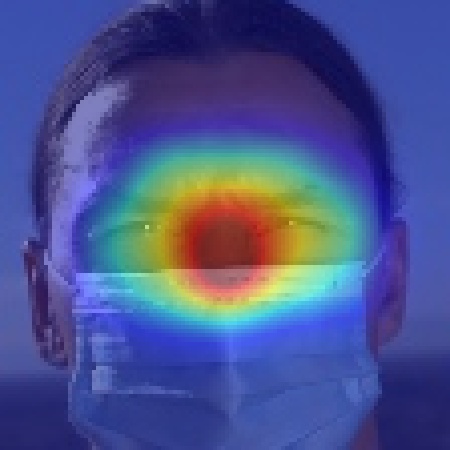}}\\
\subfloat[Upper-face Occlusion]{\includegraphics[width = 0.9in]{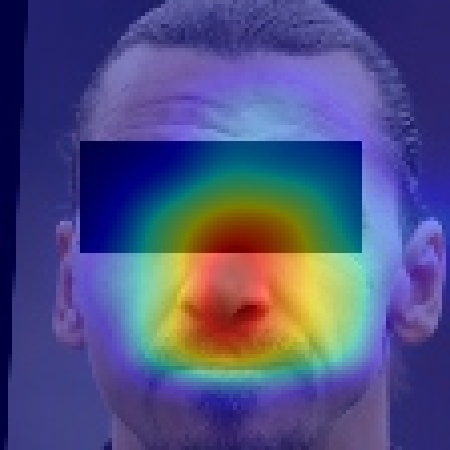}}&
\subfloat[Lower-face Occlusion]{\includegraphics[width = 0.9in]{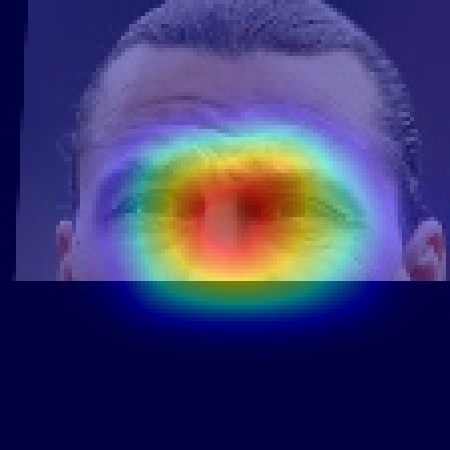}} &
\subfloat[Synthetic Mask]{\includegraphics[width = 0.9in]{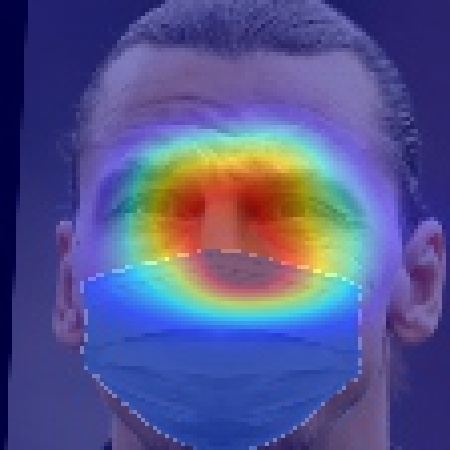}} &
\end{tabular}
\caption{The regions that VGGFace2 targets the most during embedding extraction \cite{Selvaraju_2019}. 
}
\label{fig:cam}
\end{figure}

To analyze the results further, we also visualized the regions that the deep face model focuses using Grad-CAM method \cite{Selvaraju_2019}. The obtained results are illustrated in Figure~\ref{fig:cam}. As can be seen the model mainly focuses on the inner face region, where eye and nose are contained. This is expected, since, especially, eye region is known to have a high discrimination power. However, as the models learn from the data the highly discriminative parts and focuses on these, when they are occluded they suffer from a performance loss. Therefore, while developing an occlusion-robust deep face recognition system, this fact has to be taken into account.

\begin{table}[]
\centering
\makebox[\textwidth]{%
\begin{tabular}{cccccc}
\hline
\multicolumn{1}{l}{\textbf{}} & \textbf{Arcface-100} & \textbf{Arcface-50} & \textbf{Arcface-34} & \textbf{MobilFaceNet} & \textbf{VGGFace2} \\ \hline
\textbf{Occlusion Type}       & \textbf{Top 1}       & \textbf{Top 1}      & \textbf{Top 1}      & \textbf{Top 1}        & \textbf{Top 1}    \\
\hline
\textbf{No occlusion}         & 99.61\%              & 99.33\%             & 99.39\%             & 99.06\%               & 99.28\%           \\
\textbf{Wearing mask}         & 85.34\%              & 76.08\%             & 73.71\%             & 70.04\%               & 79.31\%           \\
\textbf{Upper occlusion}      & 98.39\%              & 96.89\%             & 96.67\%             & 89.78\%               & 89.28\%           \\
\textbf{Lower occlusion}      & 98.83\%              & 97.67\%             & 97.11\%             & 92.06\%               & 93.94\%           \\
\textbf{Synthetic masked}     & 99.00\%              & 97.78\%             & 97.78\%             & 93.22\%               & 94.83\%           \\ \hline
\end{tabular}}
\caption{Face identification results using protocol 2}
\label{tab:protocol2rec}
\end{table}


\vspace*{-5mm}
\subsection{Impact of Occlusions on Face Verification}
For the sake of completeness, we also run face verification experiments using the proposed ROF dataset. The results of experiments using protocol 1 and 2 are presented in Tables \ref{tab:protocol1ver} and \ref{tab:protocol2ver}, respectively. Similar observations can be derived from these experiments: ArcFace is found to be more robust to synthetic occlusions. The EER increases significantly when the models are tested on real world occluded faces.

\begin{table}[]
\centering
\makebox[\textwidth]{%
\begin{tabular}{cccccc}
\hline
\multicolumn{1}{l}{\textbf{}} & \multicolumn{1}{l}{\textbf{Arcface-100}} & \multicolumn{1}{l}{\textbf{Arcface-50}} & \multicolumn{1}{l}{\textbf{Arcface-34}} & \multicolumn{1}{l}{\textbf{MobilFaceNet}} & \multicolumn{1}{l}{\textbf{VGGFace2}} \\ 
\hline
\textbf{Occlusion Type}       & \textbf{EER}                                      & \textbf{EER}                                     & \textbf{EER}                                     & \textbf{EER}                              & \textbf{EER}                                     \\
\hline
\textbf{No occlusion}                & 0.011                                          & 0.014                                         & 0.014                                         & 0.021                                  & 0.017                                         \\
\textbf{Wearing sunglasses}           & 0.088                                          & 0.091                                         & 0.095                                         & 0.106                                   & 0.096                                         \\
\textbf{Upper occlusion}      & 0.024                                          & 0.036                                         & 0.041                                          & 0.073                                  & 0.076                                         \\
\textbf{Lower occlusion}      & 0.021                                          & 0.028                                         & 0.033                                         & 0.054                                  & 0.053                                         \\
\textbf{Synthetic masked}      & 0.025                                          & 0.033                                         & 0.036                                         & 0.068                                  & 0.059                                          \\ \hline
\end{tabular}}
\caption{Face verification results using protocol 1}
\label{tab:protocol1ver}
\end{table}

\begin{table}[]
\centering
\makebox[\textwidth]{%
\begin{tabular}{cccccc}
\hline
\multicolumn{1}{l}{\textbf{}} & \textbf{Arcface-100} & \textbf{Arcface-50} & \textbf{Arcface-34} & \textbf{MobilFaceNet} & \textbf{VGGFace2} \\ \hline
\textbf{Occlusion Type}       & \textbf{EER}                  & \textbf{EER}                 & \textbf{EER}                 & \textbf{EER}          & \textbf{EER}                 \\
\hline
\textbf{No occlusion}                & 0.013                      & 0.019                     & 0.021                     & 0.024              & 0.017                     \\
\textbf{Wearing mask}               & 0.083                      & 0.119                     & 0.119                     & 0.119              & 0.082                     \\
\textbf{Upper occlusion}      & 0.031                      & 0.035                     & 0.043                     & 0.067              & 0.058                     \\
\textbf{Lower occlusion}      & 0.019                      & 0.036                     & 0.038                     & 0.054              & 0.045                     \\
\textbf{Synthetic masked}      & 0.028                      & 0.036                     & 0.043                     & 0.074              & 0.049                     \\
 \hline
\end{tabular}}
\caption{Face verification results using protocol 2}
\label{tab:protocol2ver}
\end{table}

\vspace*{-5mm}
\section{Conclusion}

In this study, we present a real-world occluded face dataset and explore the effects of occlusion on the state-of-the-art face recognition methods' performance. We have shown that synthetically generated occlusions do not reflect the nature of the real-world occlusions. We have observed significant performance drops when deep face models are tested on real world occluded faces that contain masks or sunglasses. Visualization of the results indicate that the deep face models mainly focus on the inner face region. Therefore, the models experience a performance loss, when this region is occluded. For our future work, we aim to expand the collected dataset and develop an occlusion-robust deep face recognition system by benefiting from the findings of this work.

\vspace*{-5mm}
\section{Acknowledgement}

This study is supported by the Istanbul Technical University Research Fund, ITU BAP, project no. 42547 and by the Scientific and Technological Research Council of Turkey (TUBITAK) project no. 120N011.

\vspace*{-5mm}
\bibliography{lniguide}

\end{document}